\documentclass[runningheads,final]{IOS-Book-Article}
\usepackage[utf8]{inputenc}
\usepackage[T1]{fontenc}
\usepackage{ifdraft}
\usepackage{graphicx}
\usepackage{grffile}
\usepackage{longtable}
\usepackage{rotating}
\usepackage{amsmath}
\usepackage{textcomp}
\usepackage{amssymb}
\usepackage{amsthm}
\usepackage[small]{caption}%
\usepackage{subcaption}
\usepackage{capt-of}
\usepackage{booktabs,placeins} %
\usepackage[inline]{enumitem} %
\usepackage{soul}
\usepackage{tikz,pgfplots}
\usetikzlibrary{arrows,shapes,positioning}
\makeatletter
\@ifclassloaded{report}
{\newtheorem{theorem}{Theorem}[chapter]}
{\newtheorem{theorem}{Theorem}}
\makeatother

\theoremstyle{definition}

\newtheorem{example}[theorem]{Example}
\newcommand{\defemph}[1]{\emph{#1}}

\newcommand{\gpp}[2][]{\todo[author=Guilherme, caption={}, #1]{#2}}
\if@todonotes@disabled

\else

\fi

\newcommand{\gppnew}[1]{\textcolor{olive}{#1}}

\newcommand{\gppold}[1]{{\color{gray}$\times$ #1}}

\newcommand{\ftdel}[1]{\textcolor{red}{$\times$ #1}}
\newcommand{\ftold}[1]{\FTdel{#1}}

\ifoptionfinal{
\renewcommand{\gppnew}[1]{#1}
\renewcommand{\gppold}[1]{} %
\renewcommand{\ftdel}[1]{}
\renewcommand{\ftold}[1]{}
}

\newcommand{\refobjective}[1]{\hyperref[objective#1]{Objective \ref*{objective#1}}}
\newcommand{\refobjectives}[2]{\hyperref[objective#1]{Objectives \ref*{objective#1}} \hyperref[objective#2]{and \ref*{objective#2}}}

\if@todonotes@disabled
\else
\fi

\newif\ifincludeincoherent
\includeincoherenttrue

\newif\ifincludeproofs
\includeproofsfalse

\newif\ifincludeNN              %
\includeNNfalse

\newcommand{\citet}{\cite}
\newcommand{\citeauthor}{\cite}
\newcommand{\citep}{\cite}

\renewcommand{\phi}{\varphi} %

\renewcommand{\emptyset}{\varnothing}

\newcommand{\pgeq}{\succeq} %
\newcommand{\pg}{\succ}

\newcommand{\pleq}{\preceq}
\newcommand{\pl}{\prec}

\newcommand{\case}[2]{{\small \mbox{\ensuremath{(#1, #2)}}}}

\newcommand{\default}{\delta}
\newcommand{\defcharac}{\ensuremath{\default_C}}
\newcommand{\defoutcome}{\ensuremath{\default_o}}
\newcommand{\nondefoutcome}{\bar{\defoutcome}}
\newcommand{\defcase}{\case{\defcharac}{\defoutcome}}

\newcommand{\newcasearg}[1][N]{\case{{#1}_C}{?}}
\newcommand{\newcasecharac}{{N_C}}

\newcommand{\casei}{\ensuremath{\alpha}}     %
\newcommand{\caseii}{\ensuremath{\beta}}     %
\newcommand{\caseiii}{\ensuremath{\gamma}}
\newcommand{\caseiv}{\ensuremath{\epsilon}}
\newcommand{\casev}{\ensuremath{\eta}}

\tikzset{attack/.style={-latex}}

\def\myAF{\ensuremath{(\Args,\attacks)}}
\def\myAFalone{\ensuremath{(\Args',\attacks')}}
\def\Args{\ensuremath{\mathit{Args}}}
\def\attacks{\ensuremath{\leadsto}}
\def\groundext{\mathbb{G}}
\def\arga{\ensuremath{\alpha}}
\def\argb{\ensuremath{\beta}}
\def\argc{\ensuremath{\gamma}}

\def\coherent{coherent}

\newcommand{\aaD}[1][D]{\ensuremath{AF_{\pgeq}(#1)}} %
\newcommand{\aaDN}[1][D]{\ensuremath{AF_{\pgeq}(#1,\newcasecharac)}}

\mathchardef\mhyphen="2D

\newcommand{\aacbr}{\ensuremath{{AA\mhyphen CBR}}} %

\newcommand{\paacbr}{\ensuremath{\aacbr_{\pgeq}}}

\newcommand{\oaacbr}{\ensuremath{\aacbr_{\supseteq}}} %
\newcommand{\caacbr}{\ensuremath{c{\paacbr}}} %

\newcommand{\welfdata}{\texttt{WelfareFailMany}}

\usepackage[textwidth=15mm,textsize=footnotesize,obeyFinal]{todonotes}
\usepackage{times}

\usepackage{array}   %
\newcolumntype{C}{>{$}c<{$}}

\begin{document}
\label{sec:orgfd9ec34}
  \begin{frontmatter}
  \title{Technical Report on the Learning of Case Relevance in Case-Based Reasoning with Abstract Argumentation}
  \runningtitle{Learning Case Relevance in Case-Based Reasoning with Abstract Argumentation}
  \gpp{max 10 pages with references}

  \author{\fnms{Guilherme} \snm{Paulino-Passos}\orcid{0000-0003-3089-1660}%
  \thanks{Corresponding Author: Guilherme Paulino-Passos, g.passos18@imperial.ac.uk.}}
and
  \author{\fnms{Francesca} \snm{Toni} \orcid{0000-0001-8194-1459}}
  \address{Imperial College London, Department of Computing, London, United Kingdom}
  \runningauthor{G. Paulino-Passos and F. Toni}

  \begin{abstract}
Case-based reasoning is known to play an important role in several legal settings. In this paper we focus on a recent approach to case-based reasoning, supported by an instantiation of abstract argumentation whereby arguments represent cases and attack between arguments results from outcome disagreement between cases and a notion of relevance. %
We explore how relevance can be learnt automatically in practice with the help of decision trees, and explore the combination of case-based reasoning with abstract argumentation (\aacbr{}) and learning of case relevance for prediction in legal settings. Specifically, we show that, for two legal datasets, \aacbr\ and decision-tree-based learning of case relevance perform competitively in comparison with decision trees%
.
We also show that \aacbr{} with decision-tree-based learning of case relevance results in a more compact %
representation than their decision tree counterparts, which could be beneficial for obtaining cognitively tractable explanations.
  \end{abstract}


  \end{frontmatter}

\section{Introduction}
\label{sec:org242fba2}

     Case-based reasoning (CBR) is a methodology in which concrete past occasions are directly used as sources of knowledge and solutions for new situations~\citep{%
     DBLP:books/daglib/0032926}.
     It has been %
     studied in AI and Law since its inception%
     , leading to foundational contributions~\citep{%
     Rissland2005CasebasedRA}. %
     This is a not a surprise, given the centrality of the use of cases for determining the law in Common Law systems, although not exclusively~\cite{lewis2021}.
     
In this paper we focus on 
recent approaches to CBR~\citep{DBLP:conf/kr/CyrasST16, %
     dear-2020, DBLP:conf/kr/Paulino-PassosT21, Prakken2022ATM} using argumentation~\citep%
     {prakken-overview%
     }.
     Argumentation itself %
     has a long history in AI and Law, and its use to support CBR has been shown to pave the way towards novel forms of explanations for the outcomes of CBR, including via arbitrated dispute trees~\citep{DBLP:journals/eswa/CyrasBGTDTGH19, DBLP:conf/ijcai/Cyras0ABT21}. %
Specifically, we focus on the 
\aacbr{} approach~\citep{DBLP:conf/kr/CyrasST16, %
dear-2020, DBLP:conf/kr/Paulino-PassosT21}, where arguments correspond to cases %
and attacks between arguments result from outcome disagreement between cases and \emph{relevance} between cases, guided by a partial order over cases capturing some notion of specificity%
. 
     Originally~\citep{DBLP:conf/kr/CyrasST16}, \aacbr{} expects a representation of cases in terms of %
     sets of  \emph{manually engineered binary} features %
      and %
      the partial order is defined via the subset relation%
      . This expectation is a restriction for applicability.
   While previous work has generalised %
   beyond binary features in order to support different applications~\cite{dear-2020}, a systematic generalisation to tabular datasets, including categorical and continuous data, is still missing. This is essential for applying \aacbr{} to %
   \gppnew{realistic} datasets, including legal ones, to realise the original inspiration from legal reasoning for \aacbr. While some form of binarisation can be applied, there is no guarantee that a naïve binarisation would result in good performance. %
   In this work we close %
   this gap, focusing on applying \aacbr\ to possibly non-binary tabular data from legal settings.

   Specifically, our first contribution is a general method for %
   applying \aacbr\ to any tabular data by extracting binary features from decision trees~\cite{DBLP:books/wa/BreimanFOS84} %
   when learning for the final task. Our second contribution is showing that this method is competitive with decision trees on two legal datasets: COMPAS~\cite{propublica-compas} and a simulated legal dataset~\cite{DBLP:journals/argcom/StegingRVB23}  %
   for welfare benefit. Finally, as a third contribution, we show that\gpp{revise according to actual final results} our method creates smaller models %
   (i.e. with a smaller number of nodes), %
   leading to potentially more cognitively tractable explanations (i.e. decision trees and rules drawn from them on one hand, and argumentation frameworks and arbitrated dispute trees on the other).

\gppold{
Explainability is a fundamental requirement in AI and Law.
In particular, any model of legal reasoning must be amenable to explanation, which is required not only of AI systems, but of legal decision-makers in general.
Current state of the art methods in many AI tasks is data-driven methodologies (machine learning)  whether machine learning-base
... \aacbr{} has been developed ...~\citep{DBLP:conf/kr/CyrasST16,dear-2020} ..

Important aspects for any concrete implementation of \aacbr{} are the representation of the data (to which we refer as characterisation) and, crucially, the partial order \(\pleq\) on the set of characterisations, which until now we have assumed given. This partial order captures the idea of relevance of past cases for a new case, as well as of specificity among past cases%
.

In this paper, we will discuss ways of defining this partial order. In particular, we are interested in learning the partial order from data. Learning the partial order allows \aacbr{} to a wider scope of problems, instead of requiring a hand-constructed order. This carries three desiderata: 1) that the learned partial order results in a good performance; 2) that it is human interpretable.
We here present a method for learning a partial order from tabular data using features extracted from decision trees trained on this same data. %
. We show empirical results via%
experiments are carried out on the the COMPAS dataset~\citep{propublica-compas}, a dataset on prediction of recidivism, that is, on prediction of whether a criminal will re-offend~\citep{2016COMPASRS}, as well as on the simulated legal data of...
}

\section{Background}
\label{sec:background}
\paragraph{Abstract Argumentation frameworks (AFs).}
\label{sec:orge6fb14a}
An \emph{AF}~\cite{Dung:95} is a pair $\myAF$,
where $\Args$ is a set (of \emph{arguments}) 
and $\attacks$$\subseteq \!\!\Args\!\times\!\Args$ is a binary relation (of \emph{attack}) on $\Args$. 
For $\arga, \argb \!\in \!\Args$, if $\arga \!\attacks \!\argb$, 
then we say that 
$\arga$ \emph{attacks} $\argb$%
. 
$E \!\subseteq \!\Args$ \emph{defends} %
$\arga \!\in \!\Args$ 
 if for all $\argb \!\attacks \!\arga $ there exists 
 $\argc \!\in \!E$ such that $\argc \!\attacks \!\argb$. 
\label{defn:semantics} 
The \emph{grounded extension} of $\myAF$ %
is $\groundext \!= \!\bigcup_{i \geqslant 0} G_i$, 
where $G_0$ is the set of all unattacked arguments, 
and $\forall i \!\geqslant \!0$, $G_{i+1}$ is the set of arguments that $G_i$ defends.

\paragraph{Abstract Argumentation for Case-Based Reasoning (\aacbr).} 
\label{sec:orgdda69ea}
We use the \paacbr{} presentation %
from
\citet{DBLP:conf/kr/Paulino-PassosT21}%
.
Let $X$ be a set of \emph{characterisations}, equipped with partial order $\pleq$%
.
Let $Y \!= \!\{\defoutcome,\nondefoutcome\}$ be a set of \emph{outcomes}, with $\defoutcome$ the \emph{default outcome}. We discriminate a particular element $\defcharac \!\in \!X$ 
such that $\defcharac$ is the $\pleq$-minimum element of $X$
and define the \emph{default argument} $\defcase \!\in \!X \!\times \!Y$.
A \emph{casebase} (aka \emph{dataset}) $D$ is a finite %
$D \!\subseteq \!X \!\times \!Y$, consisting of
\emph{past cases} (aka \emph{labelled examples}) $\alpha\in D$ is 
of the form $\case{\alpha_{C}}{\alpha_{o}}$ for $\alpha_{C}\!\in \!X$, $\alpha_{o}\!\in \!Y$. Instead, a \emph{new case} (aka \emph{unlabelled example}) is of the form $\newcasearg$ for $\newcasecharac \!\in \!X$. %

The partial order $\pleq$ defines a notion of \emph{relevance} $\sim$ between characterisations, where $x_1 \! \sim{} \! x_2$ iff $x_2 \!\pleq\! x_1$. This notion and crucially \emph{irrelevance} (defined as $\not \sim$) are used to compare new and past cases as well as two past cases.  (thus in \paacbr{} relevance is not symmetric). \gppnew{The idea is that the partial order $\pleq$ captures \emph{specificity} between cases, %
  and that
  the outcome for a new case depends only on past cases than which the new case is more specific.}
For simplicity%
        , we %
        extend the definition of $\pgeq$ (and $\nsim$) to %
        cases by setting $\case{\alpha_c}{\alpha_o} \pgeq \case{\beta_c}{\beta_o}$ iff $\alpha_c \pgeq \beta_c$ (and $\newcasearg \!\!\not \sim \!\!(\beta_C,\!\beta_o)$ iff $\newcasecharac\ \!\!\not \sim \!\!\beta_C$). \gpp{see if needed} 
For characterisations as sets of binary features, as in \citep{DBLP:conf/kr/CyrasST16}, $\pgeq= \supseteq$ captures specificity.\todo{mention relevance somehow? more important than specificity, for how we have presented matters}

\gppold{\paacbr{} maps a (finite) {\emph{dataset}} \(D\) of {{\em examples}} (each labelled with one of two {\em outcomes}) and an {\em {unlabelled example}} (with unknown outcome) into an AF. The dataset may be understood as a {\em casebase}, the labelled examples as {\em past cases} and the unlabelled example as a {\em new case}: we will use these terminologies interchangeably throughout.
Cases are represented by a \emph{characterisation} and we call $X$ the set of all characterisations%
. One of the two outcomes is selected up-front as the \emph{default outcome} ($\defoutcome$, the non-default being $\nondefoutcome$, and $Y \!= \!\{\defoutcome,\nondefoutcome\}$).
Finally, \paacbr\ assumes that the set of characterisations of (past and new) cases is equipped with a partial order {$\pleq$} (whereby $\arga \!\prec \!\argb$ {holds if $\arga \!\pleq \!\argb$ and $\arga \!\neq \!\argb$ and} is read ``$\arga$ is less \emph{specific} than $\argb$'') and with a relation $\not \sim$  (whereby $\arga \!\not\sim \!\argb$ is read as ``$\argb$ is {\em irrelevant} to $\arga$''), which here we restrict to be defined as $\argb \!\not\pleq \!\arga$.}

  \label{dear-miner}
\gppold{  Let $X$ be a set of \emph{characterisations}, equipped with partial order $\pl$ and binary relation $\not\sim$.  Let $Y \!= \!\{\defoutcome,\nondefoutcome\}$ be the set of (all possible) \emph{outcomes}, with $\defoutcome$ the {\em default outcome}.  
Then, a {\em casebase} $D$ is a finite set such that  $D \!\subseteq \!X \!\times \!Y$
        (thus a {\em past case} $\alpha\in D$   is of the form $\case{\alpha_{C}}{\alpha_{o}}$ for $\alpha_{C}\!\in \!X$, $\alpha_{o}\!\in \!Y$)
        and a {\em new case}  is of the form $\newcasearg$  for $\newcasecharac \!\in \!X$.
        {We also discriminate a particular element $\defcharac \!\in \!X$ and define the \emph{default argument} $\defcase \!\in \!X \!\times \!Y$.}}
        A casebase $D$ is {\em \coherent} iff there are no two cases  $\case{\alpha_{C}}{\alpha_{o}},\case{\beta_{C}}{\beta_{o}}\in D$ such that $\alpha_{C} = \beta_{C}$ but $\alpha_{o} \neq \beta_{o}$, and it is \emph{in\coherent} otherwise. \gpp{consider removing, if not needed}

  \gpp{Double check defs with thesis}

  The \emph{AF mined from a dataset $D$ and a new case $\newcasearg$}, given default argument $\defcase$, is $\myAF$ (referred to as \aaDN\ later), in which: 
  
  \begin{enumerate}[label=(\roman*)] %
    \vspace{-.5em}
  \item $\Args=  D \cup \{\defcase\} \cup \{\newcasearg\}$;    
  \item  for $(\alpha_C, \alpha_o) \in D$, $(\beta_C, \beta_o) \in D \cup \{ \defcase \}$, it holds that $(\alpha_C, \alpha_o) \attacks (\beta_C, \beta_o)$ iff

    (a) $\alpha_o \neq \beta_o$,
    \quad (b) {$\alpha_C  \pgeq \beta_C$, \quad and}

    (c) $\nexists (\gamma_C, \gamma_o) \in D\cup \{ \defcase \}$ with $\alpha_C \pg \gamma_C \pg \beta_C$ and {$\gamma_o = \alpha_o$}; %
                
  \item  for $(\beta_C, \!\beta_o) \!\!\in \!\!D \cup \!\{ \!\defcase \!\}$, it holds that $\case{\newcasecharac}{?} \!\!\attacks \!\!(\beta_C, \!\beta_o)$ iff 
    $\newcasearg \!\!\not \sim \!\!(\beta_C,\!\beta_o)$.%
    \vspace{-.5em} %
  \end{enumerate}
                Let $\groundext$ be the grounded extension of %
        \aaDN.  
  Then, the \defemph{outcome} \defemph{for $\newcasecharac$} is $\defoutcome$ if $\defcase$ is in $\groundext$, and $\nondefoutcome$ otherwise. 
  
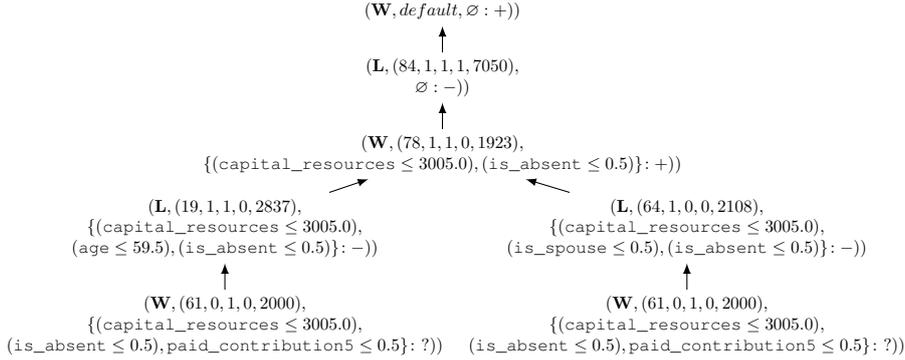
\begin{figure}[hbt]
  \centering
\begin{tikzpicture}[>=latex,line join=bevel, scale=1, every node/.style={scale=.65}, %
  ]

  \node (default) at (391.0bp,354.0bp) [align=center] {$(\mathbf{W}, default, \emptyset{}: +))$};
  \node (c1) at (391.0bp,285.5bp) [align=center, below=1em of default] {$(\mathbf{L}, (84,1,1,1,7050),$\\$ \emptyset{}: -))$};
  \node (c2) at (391.0bp,208.5bp) [align=center, below=1em of c1] {$(\mathbf{W}, (78,1,1,0,1923),$\\$ \{(\texttt{capital\_resources} \leq 3005.0), (\texttt{is\_absent} \leq 0.5)\}\!: +))$};
  \node (c3a) at (225.0bp,123.0bp) [align=center,below left=.5em and -7em of c2] {$(\mathbf{L}, (19,1,1,0,2837),$\\$ \{(\texttt{capital\_resources} \leq 3005.0),$\\$ (\texttt{age} \leq 59.5), (\texttt{is\_absent} \leq 0.5)\}\!: -))$};
  \node (c3b) at (588.0bp,123.0bp) [align=center,below right=.5em and -7em of c2] {$(\mathbf{L}, (64,1,0,0,2108),$\\$ \{(\texttt{capital\_resources} \leq 3005.0),$\\$ (\texttt{is\_spouse} \leq 0.5), (\texttt{is\_absent} \leq 0.5)\}\!: -))$};
    \node (new) at (618.0bp,29.0bp) [align=center,below=1em of c3b] {$(\mathbf{W}, (61,0,1,0,2000),$\\$ \{(\texttt{capital\_resources} \leq 3005.0),$\\$(\texttt{is\_absent} \leq 0.5), \texttt{paid\_contribution5} \leq 0.5\}\!:\!~?))$};
  \node (new2) at (166.0bp,29.0bp) [align=center,below=1em of c3a] {$(\mathbf{W}, (61,0,1,0,2000),$\\$ \{(\texttt{capital\_resources} \leq 3005.0),$\\$ (\texttt{is\_absent} \leq 0.5), \texttt{paid\_contribution5} \leq 0.5\}\!:\!~?))$};
  \draw [->] (c1) -- (default);
  \draw [->] (c2) -- (c1);
  \draw [->] (c3a) -- (c2);
  \draw [->] (c3b) -- (c2);
  \draw [->] (new) -- (c3b);
  \draw [->] (new2) -- (c3a);
\end{tikzpicture}
\caption{ADT originated from a \paacbr{} model trained on the Welfare dataset. Nodes are labelled as follows: whether winning ($\mathbf{W}$) or losing ($\mathbf{L}$); feature values of the case for: \texttt{age}, \texttt{paid\_contribution5}, \texttt{is\_spouse}, \texttt{is\_absent}, \texttt{capital\_resources}; the set of features extracted from the decision tree splits; the case outcome (``$+$'' for eligible, ``$-$'' for ineligible, and ``$?$'' for new case).}

  \label{fig:adt}
\end{figure}

      In the remainder, we will also use the notion of
      \emph{AF mined from a dataset $D$ alone} (referred to as \aaD), amounting to $\myAFalone$ 
        with 
  $\Args'=  \Args \setminus \{\newcasearg\}$ and
        $\attacks' =\attacks \cap (\Args'\times \Args')$ where
        \aaDN=$\myAF$.
Besides experimenting with (learnt instances of) \paacbr, we will also experiment with the \emph{cumulative} version thereof (\caacbr{}) from \citep{DBLP:conf/kr/Paulino-PassosT21} (definition omitted for lack of space).
\gppnew{Finally, we will use arbitrated dispute trees (ADT) as explanations~\cite{DBLP:journals/eswa/CyrasBGTDTGH19}. Intuitively, ADTs capture a debate in which, if the classification is the default outcome, then the winner successfully defends every attacker of the default argument, and otherwise the winner presents a successful attack to the default argument (example in Figure~\ref{fig:adt})
  .}

\section{Learning Relevance}
\label{sec:org16ae8d4}

\gppnew{Learning relevance in \paacbr{} amounts to learning the partial order $\pgeq$, which represents specificity.
We can think of it in terms of the 4R cycle of CBR: retrieve, reuse, revise, and retain~\cite{DBLP:books/daglib/0032926}. Retrieval is determined by (ir)relevance, which in turn is determined by the partial order. Reuse of solutions depend on the structure of the AF mined from the data, which again depends exclusively on case characterisation and on the partial order. Assuming revision to be external (e.g. by human or environmental feedback), then finally retaining is also dependent on the partial order, since retaining is nothing more than adding into the mined AF, which is also determined by the partial order.}

\begin{figure}[tb]
       \begin{center}
         \includegraphics[width=1.0\linewidth]{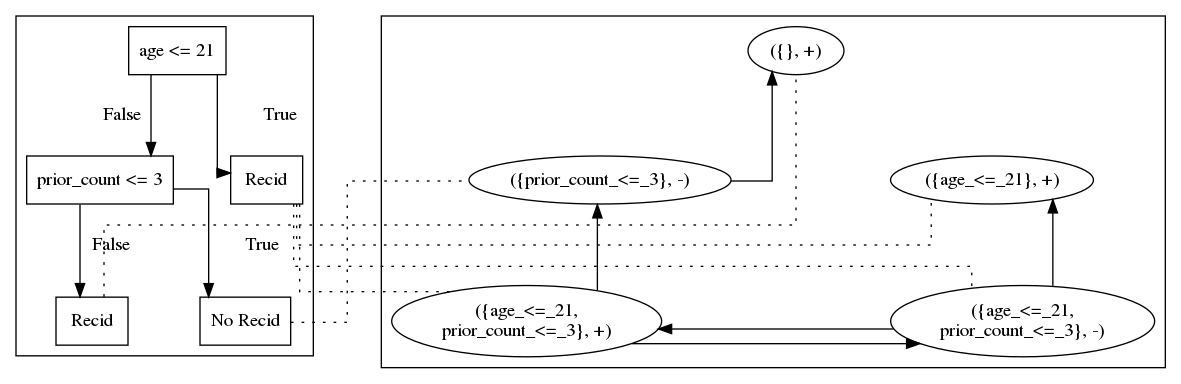}
         \caption{On the left, decision tree learnt from the dataset in Example~\ref{ex-learn-ae-dt}%
         . On the right, 
         \aaD{} for $D$ drawn from the splits in the decision tree. %
         Dotted lines indicate %
         correspondence between a leaf node on the left and a case on the right.}

         \label{fig-learn-ae-dt}
     \end{center}
     \end{figure}

Here we use decision tree learning, a classic and widely used machine learning method, to extract characterisations suitable for \oaacbr{} (i.e. \paacbr\ with $\pgeq=\supseteq$) from 
tabular data, regardless of the nature of features therein. 
Specifically, %
we use %
the algorithm CART for learning decision trees, in which decision nodes are greedily created choosing the feature and the split threshold which minimises some loss function: those splits result in a binary divide between examples which are below the split threshold and the ones above; each split can then be seen as a binary feature, and each example can be characterised 
 simply as a set of binary features, that is, the set of split rules for which the example is below the threshold. %
\gppnew{Specificity here then simply means having all (binary) features of another case.}
The following example provides a simple illustration of our method.
 \begin{example}
 \label{ex-learn-ae-dt}
 Consider the dataset %
 with the following labelled examples:
 {\footnotesize \begin{align*}
 \casei &= \case{(\text{age} = 20, \text{prior\_count} = 2)}{\texttt{+}}, & \caseiii &= \case{(\text{age} = 35, \text{prior\_count} = 7)}{\texttt{+}}, \\
   \caseii &= \case{(\text{age} = 30, \text{prior\_count} = 1)}{\texttt{-}}, & \caseiv &= \case{(\text{age} = 19, \text{prior\_count} = 1)}{\texttt{-}}, \\   
 \casev &= \case{(\text{age} = 19, \text{prior\_count} = 10)}{\texttt{+}}
 \end{align*}}
 Assume as well the decision tree on the left of Figure~\ref{fig-learn-ae-dt} was trained on this dataset. 
 Assume further that 
 the default outcome for \paacbr{}is $+$, reflecting that the majority of the dataset has the output \texttt{recid}. 
 Then, on the right of Figure~\ref{fig-learn-ae-dt}  we show \aaD, the AF mined from the dataset $D$ obtained from the original dataset by binarising features%
 : each example in %
 the original dataset is represented by the split tests for which it is evaluated \texttt{true}. %
 The resulting $D$ is then used as a casebase for \oaacbr{}. 
 Notice how there is no bijection between leaves in the decision tree and cases in the AF. The correspondence is one to many, since examples falling into the same leaf may correspond to different cases in the AF. For example, see how the rightmost leaf corresponds to $\case{\{\texttt{age\_<=\_21}\}}{+}$ (via case $\casev$), to $\case{\{\texttt{age\_<=\_21}, \texttt{prior\_count\_<=\_3}\}}{+}$ (via  case $\casei$), and finally to $\case{\{\texttt{age\_<=\_21}, \texttt{prior\_count\_<=\_3}\}}{-}$ (via case $\caseiv$).
 Notice also 
 that there is no case $\case{\{\}}{-}$ resulting from the splits, since the only cases with no features has the \texttt{recid} outcome (had there been such a case then it would have been a better choice for the default argument).
         Also, when multiple cases would have the same characterisation but different outcomes, either an incoherence is generated, or it is avoided via preprocessing. Here we illustrate an incoherence occurring.
 \end{example}

     \todo{bring back discussion on spikes, if important for the experiments later on}
     \gpp{it matters for number of nodes, but not crucial. but for reproducing would be nice at least mentioning we are removing them.}
     
\section{Experiments}
\label{sec:experiments}

We train decision trees with pre-pruning, that is, limiting maximum depth and maximum number of leaf nodes for the decision tree as a regularisation, where the best maximum vales are hyperparameters chosen by cross-validation. We used the following set of values: for maximum depth, varying from 3 to 13, in a step of 2; for maximum number of leaf nodes, from 4 to 512, in geometric progression of ratio 2. Nodes are greedily created in a best-first search fashion\gppnew{, using Gini impurity as the criterion}.
\gpp{! moved here to Experiments, before COMPAS, but OK to move to before experiments since part of the approach}%
We evaluate three approaches for the problem of incoherence:
\begin{enumerate*}
\item \texttt{keep}: to keep the incoherence and letting each model deal with this in their own ways;
\item \texttt{removal}: to remove every incoherent pair of cases;
\item \texttt{majority}: for each characterisation in the resulting transformation, count the number of training examples corresponding to each output and select the majority output as the outcome for the (now unique) case.
\end{enumerate*}

\subsection{COMPAS Dataset}
\label{sec:compas}
The first dataset we use is COMPAS, which contains predicted scores of recidivism and data of actual (measured) recidivism~\citep{propublica-compas}. The COMPAS dataset is based on a proprietary prediction model for recidivism widely used in the U.S~\citep{propublica-compas} and it has been the focus of much work on algorithmic biases and impact of technology in the justice system. Our goal is not evaluating the COMPAS algorithm or discuss its fairness~\citep{propublica-compas}%
, but simply using this dataset as a way of evaluating our methodologies on learning relevance for CBR in a legally relevant scenario. %
This should not be seen as results of a ready-to-deploy system or which allow clear conclusions from a criminal justice point of view.

We use the two-year recidivism dataset and apply the original filtering strategy for missing data%
, resulting in 6172 entries. %
This is a tabular dataset, each corresponding to a defendant who was screened by COMPAS before trial, and given a COMPAS score on risk of recidivism.
For each person, the dataset contains personal information (such as age, gender, and race), information about the current charge (such as degree of seriousness of the charge and days in which the defendant was imprisoned); criminal history and whether the defendant has reoffended%
.
\gpp{perhaps mention that possibly problematic definitions, possibly innocent defendants, including reoffending, perhaps on the ethics part}
We experimented with 4 different feature sets:
\begin{enumerate*}[label=(\Alph*)]
\item containing all features. \label{fs:all}
\item removing \texttt{age\_cat};
\item removing \texttt{age\_cat} and race;
\item removing \texttt{age\_cat}, race and gender;
\end{enumerate*}
We do so since \texttt{age\_cat} is redundant with the age feature, while  %
race and gender %
are protected features, thus we consider a kind of fairness through unawareness. In the remainder, when we refer to the dataset not specifying a feature set, we mean feature set C.

\begin{table}[tb]
  \centering
  \caption{Percentage accuracy of each \aacbr{} model and each strategy for incoherence in the casebase, aggregated over hyperparameter choice (maximum depth and maximum number of leaves in the decision tree). Results on COMPAS test set, feature set A.}
  \label{tab-learn-incoherence}
  \begin{tabular}{@{}lCCCCCC@{}}
    \toprule
    & \multicolumn{3}{C}{\paacbr} & \multicolumn{3}{C}{\caacbr} \\[0pt]
    \cmidrule(lr){2-4}\cmidrule(lr){5-7}
& \text{keep} & \text{removal} & \text{majority} & \text{keep} & \text{removal} & \text{majority}\\[0pt]
    \midrule
min&45.6 & 54.4 & 58.2 & 47.0 & 54.4 & 57.5\\[0pt]
max&55.3 & 63.9 & 68.1 & 57.8 & 61.9 & 68.1\\[0pt]
avg\textpm{}stddev&49.1\pm{}4.3 & 57.3\pm{}2.2 & 64.1\pm{}4.0 & 52.3\pm{}3.0 & 57.4\pm{}2.4 & 63.9\pm{}4.2\\[0pt]
    \bottomrule
  \end{tabular}
\end{table}

\paragraph{Results.}
Comparing the strategies for dealing with incoherence, \texttt{keep} is a weaker strategy than the other two even for \caacbr{}, which deals with incoherence directly, while \texttt{majority} is the stronger strategy. %
This strategy dominance is shown not only via on the test set directly (Table~\ref{tab-learn-incoherence}) but also over almost all hyperparameter choices (Table~\ref{tab-learn-incoherence-delta}). That is, choosing optimal hyperparameters appropriately results in using \texttt{majority}. %

\begin{table}[tbh]
  \centering
  \caption{Difference in percentage accuracy between the \texttt{removal} or \texttt{keep} strategies and the \texttt{keep} strategy for incoherence, by \aacbr{} model, aggregated over hyperparameter choice (maximum depth and maximum number of leaves in the decision tree). Aggregation is performed over the difference. Results on COMPAS test set, feature set A.}
  \label{tab-learn-incoherence-delta}
  
  \begin{tabular}{@{}lCCCC@{}}
    \toprule
    &  \multicolumn{2}{C}{\paacbr} & \multicolumn{2}{C}{\caacbr} \\[0pt]
    \cmidrule(lr){2-3}\cmidrule(lr){4-5}
    &  \text{removal} - \text{keep} & \text{majority} - \text{keep} & \text{removal} - \text{keep} & \text{majority} - \text{keep}\\[0pt]
    \midrule
    min &  2.1 & 2.9 & -0.1 & 1.8\\[0pt]
    max &  13.0 & 22.5 & 10.4 & 21.00\\[0pt]
    mean\textpm{}stddev  & 8.1\pm{}4.1 & 15.0\pm{}8.2 & 5.1\pm{}3.5 & 11.6\pm{}7.1\\[0pt]
    \bottomrule
  \end{tabular}

\end{table}
On Table~\ref{tab-learn-dt-results} we directly compare performance for each of three models: \paacbr, \caacbr, and decision trees.
We do 5-fold cross validation. In each fold, hyperparameters are selected in a inner validation step (using a single split for validation set), retrained on the entire training set of that fold, and evaluated on the test set of the fold. This is done for each model, and we report results in Table \ref{tab-learn-dt-results}, aggregated by fold.\gpp{if necessary to cut something to add acknowledges or increase figure etc, cut paragraph until now except 1st sentence}
Comparing the different feature sets, we can see that, under our method for learning relevance, \paacbr{} and \caacbr{} show comparable performance with decision trees on COMPAS.
An interesting result in this experiment is that in most cases the optimal hyperparameter choice for each of \paacbr{} and \caacbr{} resulted in both having the same classification in the test set. This suggests they may have the same decision function%
, even if they have different inner structure. %

\begin{table}[tb]
\centering
\caption{Performance measured in percentage accuracy for COMPAS, for each approach, averaged over 5-fold cross validation, with standard deviation, and using hyperparameter search by internal validation split. Reported by feature set.}
\label{tab-learn-dt-results}
\begin{tabular}{@{}lCCCC@{}}
\toprule
  {} & \multicolumn{1}{l}{Feature set A} & \multicolumn{1}{l}{Feature set B} & \multicolumn{1}{l}{Feature set C} & \multicolumn{1}{l}{Feature set D} \\
\midrule
Decision tree &        67.60\pm{}1.31&        67.60\pm{}1.31&        67.48\pm{}1.56&        67.00\pm{}1.15\\
\paacbr        &        66.32\pm{}1.20&        66.32\pm{}1.20&        66.32\pm{}1.20&        66.41\pm{}1.31\\
\caacbr        &        66.32\pm{}1.20&        66.32\pm{}1.20&        66.32\pm{}1.20&        66.41\pm{}1.31\\
\bottomrule
\end{tabular}
\end{table}

         \gpp{add around here new figs}

\subsection{Welfare Benefit Dataset}
The welfare benefit domain was originally proposed in \cite{DBLP:conf/icail/Bench-Capon93}, with the goal of having a dataset that captures conditions typically found in law. This dataset concerns the eligibility of a person for a welfare benefit to cover the expenses for visiting their spouse in the hospital. The task is binary classification of whether the person is eligible for the benefit or not.
This is a simulated dataset, not based on natural distributions, developed for evaluating rationales of machine learning models. %
Our goal is evaluating our method for learning relevance for \paacbr{} and a thoroughly evaluation of rationale is outside our scope.%
We use the available \welfdata{} dataset, containing contains 2000 cases, where 1000 are eligible cases and 1000 are ineligible.

\paragraph{Results.}
Table~\ref{tab:welfare-results} shows that \texttt{majority} is the stronger strategy also for Welfare. Interestingly, for \paacbr{} \texttt{keep} shows better performance than \texttt{removal}, that presents very high variance. By inspection of the learned models, this happened since many such learned models end up containing very few cases or even just the default case, due to the learned representation having always incoherent cases for each or many characterisations. This also suggests a higher sensibility of \caacbr{} to noise. This is shown here by the high variance of the \texttt{removal} strategy. On the other hand, \texttt{majority} has not only a higher average, but is also more stable, with a small variance. Overall, the results confirm the ones seen for COMPAS, where \texttt{majority} is a better strategy in which both \aacbr{} approaches show performance on par with decision trees.

\begin{table}[tbh]
  \centering
  \caption{Performance measured in percentage accuracy for Welfare, for each approach and each strategy for incoherence, using hyperparameter search by internal validation split. Averages over 5-fold cross-validation, with standard deviation.}
    \begin{tabular}{@{}lCCCCCCC@{}}
      \toprule
     &  \multicolumn{1}{c}{Decision Tree} & \multicolumn{3}{c}{\paacbr{}} & \multicolumn{3}{c}{\caacbr{}} \\
      \cmidrule(lr){2-2}\cmidrule(lr){3-5}\cmidrule(lr){6-8}
     &           & \text{keep} & \text{removal} & \text{majority} & \text{keep} & \text{removal} & \text{majority} \\
    \midrule
    & 99.6\pm{}0.1 & 99.3\pm{}0.6 & 90.5\pm{}18.0 & 99.5\pm{}0.4 & 82.9\pm{}18.8 & 90.5\pm{}18.0 & 99.6\pm{}0.2 \\
    \bottomrule
\end{tabular}
  \label{tab:welfare-results}
\end{table}

\subsection{Explainability}
\label{sec:explainability}
Explanations come in two forms: global explanations, which explain the behaviour of entire model over all possible inputs; and local explanations, which explain the behaviour of a particular prediction~\cite{molnar2022}.
Given that both decision trees and \paacbr{} are intrinsically interpretable models, the models themselves are subject to human inspection and can thus be evaluated as global explanations.
As for local explanations, while many are possible, we use explanations tailored for each model. For decision trees, we consider simply the decision path traversed by the classified example~\cite{molnar2022}. As for \paacbr{}, we use ADTs (§\ref{sec:background}). We choose the ADT that minimises the number of nodes by a minimax tree search algorithm.

There are no standard methodologies in the literature to evaluate explanations, with different works evaluating different aspects~\cite{molnar2022}. We here decide to use explanation size as a proxy for ease of interpretation. We do so given a known objection to the interpretability of decision trees: they are hard to interpret if they are too big~\cite{molnar2022}. All explanations that we use are in the form of graphs, and thus we can evaluate size in a uniform way.
While for decision trees the depth is commonly considered, \paacbr{} is not restricted to be a binary tree, and thus the number of nodes is no longer bound by the depth. Thus we also report the number of nodes. Finally, since ADTs contain multiple occurrences of the same cases, we also measure the number of unique nodes (which is also the size of the sub-graph of the AF corresponding to the ADT).

\paragraph{Results. }
Since the \aacbr{} models are generated from the splits in the decision and as illustrated on Figure~\ref{fig-learn-ae-dt}, a single leaf can become many nodes in \paacbr{} and \caacbr{}. While only half of the nodes of the decision tree are leafs, \aacbr{} could suffer from combinatorial explosion with many features. However, this is not what we see empirically (Table~\ref{tab:global-exp}). For COMPAS we see a $91.2\%$ reduction in of the average size for \paacbr{} and $94.2\%$ for \caacbr{}. This is subject to the high variance in decision tree size, but the \aacbr{} models show a consistent smaller size. For Welfare there is a $29.1\%$ reduction of the average size for \paacbr{} and $58.8\%$ for \caacbr{}, while the issue of variance for decision tree sizes does not occur.
This means that, for comparable accuracy, \paacbr{} and (specially) \caacbr{} can generate notably smaller models. Thus, for scenarios where an interpretable graph form of the model is required, \paacbr{} and \caacbr{} present a strong advantage over decision trees.

As for the local explanations (Table~\ref{tab:local-exp}), ADTs show a larger number of nodes than decision paths. This is expected, since ADTs require multiple occurrences of many cases (indeed, of sub-graphs) of the original AF. ADTs for \caacbr{} show comparable number of nodes to decision paths in COMPAS, but are still larger in Welfare.
The number of unique nodes is considerably smaller than decision paths for \paacbr{} and marginally so for \caacbr{} (within 1 standard deviation).
Furthermore, both \aacbr{} approaches result in a reduced depth as compared to decision paths. Thus, ADTs result in wider explanations, with multiple paths in the tree, but each of it smaller than decision paths.
Besides, it should be considered that an important difference between the \aacbr{} approaches and decision trees is that every node in \aacbr{} corresponds to at least one case in the casebase, with each node contains some counterfactual information (namely, at least what would the outcome be for an input exactly equal as the past case, but not only \cite{DBLP:conf/comma/Paulino-PassosT22}). Therefore the smaller global representations also contain more information, despite requiring a more complex computation for evaluation. This reflects into the size of the local representations, with more nodes end up being required for given a sufficient explanation. The trade-off is favourable for \aacbr{}, especially for \caacbr{}, which has ADTs of similar size to decision paths.

\begin{table}[tb]
  \centering
  \caption{Size of models, comparing number of nodes. Results for COMPAS from feature set C. Averages over 5-fold cross-validation, with standard deviation.}
    \begin{tabular}{@{}lCCCCCC@{}}
    \toprule
      & \multicolumn{3}{c}{COMPAS} & \multicolumn{3}{c}{Welfare} \\ \cmidrule(lr){2-4}\cmidrule(lr){5-7}
      & \multicolumn{1}{c}{Decision Tree} & \multicolumn{1}{c}{\paacbr{}} & \multicolumn{1}{c}{\caacbr{}}
      & \multicolumn{1}{c}{Decision Tree} & \multicolumn{1}{c}{\paacbr{}} & \multicolumn{1}{c}{\caacbr{}}\\
    \midrule
& 143.0\pm{}184.9 & 12.6\pm{}3.1 & 8.2 \pm{} 1.6 &
11.0\pm{}0.0 & 7.8 \pm{}4.3 & 4.6 \pm{}0.5\\[0pt]
    \bottomrule
\end{tabular}
  \label{tab:global-exp}
\end{table}

\begin{table}[tb]
  \centering
  \caption{Size of local explanations, by depth, number of nodes, and number of unique nodes. For decision trees, we use decision paths, where those metrics coincide. For \paacbr{} and \caacbr{}, we use minimal ADTs, and the number of unique nodes correspond to the number of cases in the original AF. Results for COMPAS from feature set C. Averages over 5-fold cross-validation, with standard deviation.}
  \begin{tabular}{@{}lCCCCCC@{}}
    \toprule
    & \multicolumn{3}{c}{COMPAS} & \multicolumn{3}{c}{Welfare} \\ \cmidrule(lr){2-4}\cmidrule(lr){5-7}
    & \multicolumn{1}{c}{depth} & \multicolumn{1}{c}{\# nodes}& \multicolumn{1}{c}{\# unique} & \multicolumn{1}{c}{depth} & \multicolumn{1}{c}{\# nodes}& \multicolumn{1}{c}{\# unique}\\
    \midrule
Decision Tree & 6.2 \pm{} 1.6 & 6.2 \pm{} 1.6& 6.2 \pm{} 1.6 & 4.2 \pm{}0.1 & 4.2 \pm{}0.1 & 4.2 \pm{}0.1 \\[0pt]
\paacbr{} & 5.6\pm{}0.3 & 11.9\pm{}1.7 & 7.9\pm{}1.1 & 3.5\pm{}0.0 & 8.1\pm{}3.9 & 5.1\pm{}1.1 \\[0pt]
    \caacbr{} & 5.9\pm{}0.4&6.1\pm{}0.4&6.0\pm{}0.3& 3.9\pm{}0.5 & 5.1\pm{}0.4 & 4.5\pm{}0.4 \\[0pt]
    \bottomrule
\end{tabular}
  \label{tab:local-exp}
\end{table}

\section{Related Work}
\label{sec:relatedwork}
   
There are different approaches in the literature for connecting CBR with machine learning methods with the goal of applying CBR. We will briefly mention some recent approaches to this issue in AI and Law.
A neural-network-based method for ascribing factors from natural language facts has been proposed in \cite{DBLP:conf/jurix/MumfordAB22}. Those factors come from a hierarchy of factors specified with ADFs~\citep{AlAbdulkarim2015FactorsIA}.        
In \cite{Grabmair_thesis}, a method using argument schemes with issues and quantitative value effects is proposed. Factors have effects on values with weights learned via an iterative training procedure, in which argumentation-based CBR is employed. The notion of relevance in this work is based on the argument schemes (since they define argument moves that cite precedents), which in turn depend on factors, issues, and values.
In \cite{DBLP:conf/hhai/WoerkomGPV22} the COMPAS dataset is also subject of CBR analysis. The authors apply the result model of precedential constraint~\cite{Horty_2012}, although for dimensions, instead of for factors. Dimensions are inferred from the data by determining a direction depending on the coefficients of a logistic regression. In terms of learning a notion of relevance in CBR, their methods essentially consider that the notion of relevance is the precedential constraint rule, and their learning is giving a total order on each feature, transforming them into dimensions. In the resulting model, they show that only $8\%$ of the dataset is consistent, and that this is caused by outliers for which opposite outcomes would be expected. %
\aacbr{} has already been used for modelling the simpler case of factor-based precedential constraint in \cite{DBLP:conf/kr/Paulino-PassosT21}, but better understanding the interplay between \aacbr{} and precedential constraint is open for future work.

Outside AI and Law, other combinations of CBR and machine learning have been proposed for adding interpretability to data-driven models \cite{DBLP:journals/air/NugentC05}%
. For instance, some ideas have been twinning a black-box with a CBR system~\cite{DBLP:conf/ijcai/KennyK19}, %
as well as neuro-symbolic methods which add a prototype layer in a neural network so inference is done by calculating similarity to past cases and predicting based on their outputs~\citep{DBLP:conf/aaai/LiLCR18%
}. The latter can be seen as methods using deep learning for automatically defining relevance in CBR.

\section{Conclusions and Future Work}
\label{sec:orgf8395e7}
\gpp[inline]{adapt/review}
In this work we presented an approach to learn case relevance, based on the partial order, for \aacbr{} from data on the case of COMPAS and Welfare Benefits, two tabular legal datasets.
We show that binary splits of decision trees learned using CART can be used as features for \aacbr{} and allow its instantiation as \oaacbr{}.

We show that this methodology is empirically successful on those datasets, having performance comparable to decision trees. While the approach may introduce noise in the dataset, we validate different strategies for processing it, establishing that using the majority strategy (most common output) has better performance for \aacbr{}.  %

We also empirically compare the size of the explanations of decision trees with the ones from \aacbr{} with relevance learned in the proposed manner. We evaluate their global explanations (in the form of the graph representation of the full models, since they are intrinsically interpretable) and their local explanations (decision paths, for decision trees; and ADTs, for \aacbr{}). We show that both \paacbr{} and particularly \caacbr{} result in considerably smaller global explanations than decision trees. Regarding local explanations, we show that the \aacbr{} approaches have larger explanations due to redundancies in the ADTs, but of smaller depth than decision paths of decision trees. The difference in explanation size notably decreases for \caacbr{}, which creates more compact models in general. %
Considering that \aacbr{} and its explanations in the form of ADTs contain a form of counterfactuality, they seem to represent model decisions more compactly than decision trees, which could be cognitively beneficial. Evaluating this difference with users and on other scenarios is a promising avenue for future work.

Other important directions for future work include comparing other forms of CBR for legal tasks \cite{Horty_2012, Grabmair_thesis, DBLP:conf/hhai/WoerkomGPV22, Prakken2022ATM} with \aacbr{} approaches, as well as investigation the question of how to learn case relevance for unstructured data, such as images and text, for allowing \aacbr{} to also be deployed in those scenarios.

\section*{Acknowledgements}
This work was supported by the European Research Council (ERC) under the European Union’s Horizon 2020 research and innovation programme (grant agreement No.101020934, ADIX), by J.P. Morgan and the Royal Academy of Engineering, UK, under the Research Chairs and Senior Research Fellowships scheme, as well by Capes (Brazil, Ph.D. Scholarship 88881.174481/2018-01).

\bibliography{thesis-refs.bib, milestones-references.bib, papers-refs.bib, jurix2023-refs.bib}

\bibliographystyle{vancouver}
\end{document}